\documentclass[conference]{IEEEtran}
\usepackage{cite}
\usepackage{amsmath,amssymb,amsfonts}
\usepackage{algorithmic}
\usepackage{graphicx}
\usepackage{textcomp}
\usepackage{xcolor}
\usepackage{booktabs}
\usepackage{array}
\usepackage{multirow}
\usepackage{url}
\usepackage[hidelinks]{hyperref}
\usepackage{tikz}
\usetikzlibrary{arrows.meta,positioning,shapes.geometric,calc,fit}
\usepackage{pgfplots}
\pgfplotsset{compat=1.18}
\hypersetup{
  pdftitle={Cost-Aware Abstention for LLM-Based IT Ticket Classification},
  pdfauthor={Sachin Gupta},
  pdfsubject={Accepted manuscript for IEEE ICTAI 2026},
  pdfkeywords={trustworthy machine learning, calibration, cost-aware abstention, selective prediction, large language models, text classification, IT service management}
}

\definecolor{rgInput}{HTML}{CFE4F2}     
\definecolor{rgTransform}{HTML}{E5E7EB} 
\definecolor{rgAuto}{HTML}{B6E2C7}      
\definecolor{rgDefer}{HTML}{FBD79A}     
\definecolor{rgStroke}{HTML}{1F3A5F}    
\definecolor{rgLine}{HTML}{0F3D5C}      
\definecolor{rgTeal}{HTML}{1F6F8B}      

\def\BibTeX{{\rm B\kern-.05em{\sc i\kern-.025em b}\kern-.08em
    T\kern-.1667em\lower.7ex\hbox{E}\kern-.125emX}}

\makeatletter
\def\ps@arxivcopyright{%
  \def\@oddhead{}\def\@evenhead{}%
  \def\@oddfoot{\hfil\parbox[t]{\textwidth}{\centering\scriptsize
    \textcopyright{} 2026 IEEE. Personal use of this material is permitted.
    Permission from IEEE must be obtained for all other uses, in any current
    or future media, including reprinting/republishing this material for
    advertising or promotional purposes, creating new collective works, for
    resale or redistribution to servers or lists, or reuse of any copyrighted
    component of this work in other works. Accepted for publication at the
    38th IEEE International Conference on Tools with Artificial Intelligence
    (ICTAI 2026).}\hfil}%
  \def\@evenfoot{\@oddfoot}}
\makeatother

\begin{document}

\title{Cost-Aware Abstention for LLM-Based\\IT Ticket Classification}

\author{\IEEEauthorblockN{Sachin Gupta}
\IEEEauthorblockA{\textit{Independent Researcher} \\
United States \\
sachinkg12@gmail.com}}

\maketitle
\thispagestyle{arxivcopyright}
\enlargethispage{-3\baselineskip}

\begin{abstract}
Wrong automatic routing of an IT support ticket can cost more than
human triage. We evaluate cost-aware abstention as an offline,
cost-conditional readiness assessment: a routing policy is selected on calibration data under
asymmetric costs and evaluated once on held-out test data. On a public
47{,}837-ticket, eight-class corpus, we compare four trained classifiers
with four large language models under zero- and
few-shot prompting. With illustrative per-ticket costs of zero for a
correct route, one for human triage, and five for a wrong route,
term-frequency--inverse-document-frequency logistic regression selects a
threshold of 0.60 and reduces test cost from 0.763 to 0.472, a 38.1\,\%
relative reduction (95\,\% bootstrap confidence interval,
35.7\,\%--40.4\,\%), while lowering wrong routes from 15.3\,\% to
3.5\,\%. In a post-hoc 16-setting analysis, calibration-selected
policies beat both always-route and always-defer in 14 settings.
The strongest prompted configuration reaches 0.477 accuracy on a
stratified 1{,}000-ticket subset, versus 0.879 for fine-tuned DistilBERT
on the same tickets. Isotonic recalibration reduces expected calibration
error but does not reliably improve ranking discrimination; one
GPT-4o-mini setting emits a single confidence value and has correctness area
under the receiver operating characteristic curve of 0.500. At the
break-even probability of 0.80, seven prompted conditions defer
every ticket, and Haiku few-shot routes 19 of 1{,}000. A
language-model-only agreement control and margin- and entropy-based
rejection scores for the logistic-regression baseline provide
nontrivial comparators. The findings are scoped
to this dataset, prompt design, model endpoints, and illustrative costs;
the work is an evaluation framework, not a new abstention algorithm or
universal deployment certification.
\end{abstract}


\section{Introduction}
\label{sec:introduction}

Enterprise IT service desks process tens of thousands of internal
tickets per month, where the ``customer'' is an employee asking
internal IT for help with hardware, access, or software (not an
external consumer). HDI/MetricNet reports the North American
average cost per voice IT service-desk ticket at US\$17.19, with
walk-up tickets averaging US\$37.52 \cite{rumburg2021cost}. Routing happens under
an asymmetric cost structure. Sending a ticket to the correct queue
is nearly free. Sending it to a human triager is moderately
costly. Sending it to the \emph{wrong} queue is much more
expensive once you account for handoff latency, double work, and
missed service levels. Accuracy or macro-F1 alone cannot represent
this asymmetry or indicate \emph{when} to trust an automated decision
\cite{el2010foundations, chow1970optimum}.

Two trends make this gap more pressing. First, large language
models (LLMs) can be prompted as zero- or few-shot text classifiers
\cite{brown2020language, wei2022chain}, potentially reducing the
need for task-specific training. Second, modern neural classifiers,
including LLMs, are known to be poorly calibrated. They often
produce high confidence values regardless of actual accuracy
\cite{guo2017calibration, kadavath2022language, tian2023just,
lin2022teaching}. Together these trends raise a concrete
operational question: \emph{should a prompted LLM be allowed to
auto-route IT tickets, and if so, under what abstention policy?}

We study this question on the public IT Service Ticket
Classification Dataset (47{,}837 tickets, eight categories). We
treat cost-aware abstention not as a way to improve a classifier but
as an empirical readiness test under stated assumptions. Predictors,
confidence estimators, rejection scores, and policies remain separate.
Our contributions are:

\begin{enumerate}
\item A comparison of four trained non-LLM classifiers
  and eight prompted LLM conditions on the same public corpus,
  including a common 1{,}000-ticket test subset.
\item A calibration analysis that reports expected calibration error
  (ECE), correctness area under the receiver operating characteristic
  curve (AUROC), and tie-aware area under the risk--coverage curve
  (AURC), plus separate
  mixed ($M{=}9$) and LLM-only ($M{=}8$) agreement pools.
\item A calibration-selected, paired-bootstrap cost analysis with
  top-probability, probability-margin, and entropy rejection scores;
  a post-hoc 16-setting cost-parameter analysis; and LLM policies
  that use a cost-derived threshold rather than test tuning.
\item A reproducible evaluation framework and per-ticket artifacts
  that expose prediction, confidence, rejection, and cost as
  separately auditable stages.
\end{enumerate}

The main observation is simple: abstention does not repair a weak
classifier. It exposes when auto-routing is cost-preferred under a
particular dataset, confidence signal, and cost model.

\emph{Positioning.} The components we combine are individually
established: Chow-style error-reject \cite{chow1970optimum},
Elkan-style cost-sensitive decision-making \cite{elkan2001foundations},
ECE-based calibration diagnostics \cite{guo2017calibration,
naeini2015obtaining}, isotonic recalibration, and
self-consistency-inspired agreement signals \cite{wang2023self}.
\emph{We introduce no new abstention algorithm.} This paper's
contribution is an empirical evaluation of deployment-relevant assumptions and its
reproducible implementation, not a universal readiness certificate.

\section{Related Work}
\label{sec:related}

\textbf{Selective prediction and abstention.}
The classical framework for rejecting uncertain inputs traces back
to Chow's error-reject trade-off \cite{chow1970optimum}. It has
been formalized for modern classifiers under the names selective
prediction \cite{el2010foundations, geifman2017selective} and
learning with abstention \cite{cortes2016learning}; a recent survey
organizes LLM abstention methods, benchmarks, and evaluation metrics
\cite{wen2025know}. The closest
line is \emph{learning to defer}, where a classifier is trained
jointly with the option to pass an instance to a human expert
\cite{madras2018predict, mozannar2020consistent}. Our training-free
confidence-threshold policy addresses the same route-or-defer action
space without joint training. We evaluate its decisions under an
explicit cost model following Elkan's cost-sensitive formulation
\cite{elkan2001foundations}.

\textbf{Calibration and overconfidence.}
Guo et al.\ \cite{guo2017calibration} documented systematic
overconfidence in deep neural classifiers and popularized Expected
Calibration Error (ECE) \cite{naeini2015obtaining} as a diagnostic.
A growing literature studies whether LLMs are well-calibrated
\cite{kadavath2022language, lin2022teaching, tian2023just,
xiong2023can}. The broad finding is that verbalized confidence
runs too high. We contribute a public-dataset measurement of LLM
ECE in an asymmetric-cost classification setting, and report
agreement-across-predictors as a confidence signal inspired
by self-consistency \cite{wang2023self}.

\textbf{LLMs for text classification.}
Off-the-shelf prompting of LLMs as classifiers
\cite{brown2020language, wei2022chain, sun2023text} is
operationally attractive, but on short, lexical-signal-heavy tasks
strong classical text classifiers such as linear SVMs
\cite{joachims1998text} remain highly competitive, as our
IT-ticket results confirm (\S\ref{sec:results-classification}). Cost-driven LLM cascades and routing lower
inference cost by escalating only hard inputs to a stronger model
\cite{chen2024frugalgpt}; we instead hold the model fixed and use
abstention to decide when \emph{not} to auto-route at all. Our
evaluation lands in the same mixed picture for the IT-ticket
domain.

\textbf{IT ticket classification.}
Automated routing of support tickets has a long applications-track
history \cite{diao2009rule, paramesh2018classifying} but is rarely
evaluated under explicit cost models or with abstention. Our
framework is structured as a deferral layer that could be integrated
into an IT service management pipeline rather than as a new classifier.

\section{Method}
\label{sec:method}

\subsection{Problem setup}
Let $\mathcal{X}$ be the space of ticket texts and $\mathcal{Y}$ a
set of $K$ disjoint downstream queues. A base classifier
$h:\mathcal{X}\to\mathcal{Y}$ emits a label. The policy-wrapped
decision $d:\mathcal{X}\to\mathcal{Y}\cup\{\bot\}$ either routes
$h(x)$ or \emph{abstains} ($\bot$); abstained tickets go to a human
triager. Each decision incurs one of three costs:
$c_{\mathrm{ok}}$ (correct auto-route),
$c_{\mathrm{triage}}$ (human triage),
and $c_{\mathrm{wrong}}$ (incorrect auto-route). The expected per-ticket
cost is
\begin{equation}
\begin{aligned}
\mathbb{E}[c] ={}& \mathbb{P}(\bot)c_{\mathrm{triage}}
 + \mathbb{P}(\text{correct},\neg\bot)c_{\mathrm{ok}} \\
&+ \mathbb{P}(\text{wrong},\neg\bot)c_{\mathrm{wrong}}.
\end{aligned}
\label{eq:cost}
\end{equation}
Our primary analysis uses $(c_{\mathrm{ok}}, c_{\mathrm{triage}},
c_{\mathrm{wrong}}) = (0, 1, 5)$. These values are
illustrative rather than calibrated to any single organization, but
they encode the qualitative ordering that motivates abstention: a
misroute is costlier than a human triage because it still incurs
the triage after a delayed hand-off plus rework, whereas a correct
auto-route is effectively free. Because the ratio is an assumption
rather than a measured quantity, every deployment conclusion is
conditional on it. Equation~\eqref{eq:breakeven} shows how local costs
determine the routing threshold; an organization must estimate its
own triage and misroute costs before use.
As a post-hoc sensitivity analysis, we also cross
$c_{\mathrm{triage}}\in\{0.5,1,1.5,2\}$ with
$c_{\mathrm{wrong}}\in\{2,5,10,20\}$ while fixing
$c_{\mathrm{ok}}=0$. The $(2,2)$ pair is an equal-cost boundary;
the other 15 pairs retain $c_{\mathrm{wrong}}>c_{\mathrm{triage}}$.

\subsection{Predictors}
We evaluate four non-LLM baselines and four prompted LLMs.

\textit{Non-LLM baselines.} TF-IDF (term frequency--inverse
document frequency) uni- and bi-gram features feed three classifiers.
Logistic regression and linear support vector machine (SVM) cap the vocabulary at 100{,}000
features; random forest caps it at 50{,}000. The headline baseline is
multiclass logistic regression (LR; L2 regularization, $C{=}1.0$).
The linear SVM head is a one-vs-rest linear support vector classifier
(SVC) wrapped in
sigmoid-calibrated probability outputs. The random forest head
uses 300 trees with default depth. Separately, the fourth baseline
fine-tunes DistilBERT (\texttt{distilbert-base-uncased}) on tokenized
ticket text from the full training split for two epochs with batch
size 32 and maximum sequence length 256. All four are fit on the
same train split.

\textit{LLM predictors.} Each LLM is queried per ticket with a
structured prompt (template \emph{v1}; the strict \emph{v2} is
used only in \S\ref{sec:results-ablation}) that lists the allowed
labels, the ticket text, and an instruction to return JSON
\verb|{"label": ..., "confidence": ...}|.
Few-shot variants prepend $k{=}3$ exemplars \emph{per class}
(24 in total, class-balanced across the eight categories), drawn
\emph{only} from the training split. The main prompts contain label
names and examples but no natural-language category definitions;
the comparison therefore characterizes these prompting conditions,
not each LLM's intrinsic capability. We use temperature 0 and
\texttt{max\_tokens}=64 for all main runs. Out-of-set labels are
tagged as invalid; malformed JSON is tagged as a parse error.
Both invalid labels and parse errors are scored as incorrect
predictions and assigned confidence 0; the threshold policies
defer such rows rather than auto-route them, while
\texttt{always\_route}, by definition, routes them and incurs
$c_{\mathrm{wrong}}$.
The four models are GPT-4o-mini \cite{openai2024gpt4omini}, Claude
Haiku 4.5 \cite{anthropic2025haiku}, Qwen2.5-7B-Instruct
\cite{qwen2024qwen25}, and Llama-3-8B-Instruct-Lite
\cite{meta2024llama3}. GPT-4o-mini and Claude Haiku are queried
through their vendor APIs.\footnote{For reproducibility, the
proprietary endpoints and query date were: OpenAI
\texttt{gpt-4o-mini} and Anthropic
\texttt{claude-haiku-4-5-20251001}, both queried on 2026-05-22. The
Anthropic model is a dated snapshot; the OpenAI alias resolves to
the provider's then-current \texttt{gpt-4o-mini} snapshot, so those
results reflect the endpoint state on that date.} The open-weights models use
Together AI endpoints \texttt{Qwen2.5-7B-Instruct-Turbo} and
\texttt{Meta-Llama-3-8B-Instruct-Lite}, provider-documented at query
time as 8-bit floating point (FP8) and 4-bit integer (INT4), respectively;
precision is absent from our saved responses and
is not independently verified. We re-run the model families through
DeepInfra (provider-documented bfloat16, or BF16) as a cross-provider check that
does not isolate quantization; our conclusions are unchanged
(\S\ref{sec:results-fullprec}).

\subsection{Confidence and rejection scores}
We evaluate probability estimates and ranking signals separately.

\textit{Model-reported.} For a non-LLM baseline this is the maximum
class-probability estimate returned by that model: multinomial
probabilities for logistic regression, sigmoid-calibrated outputs
for the SVM, tree-vote fractions for the random forest, and softmax
for DistilBERT. For LLMs it is the \verb|confidence| field parsed
from the JSON output.

\textit{Cross-predictor agreement.} Given $M$ predictors evaluated
on the same ticket, the agreement-confidence for the primary
predictor is the fraction of the $M$ predictors that emit the same
label. No held-out labels are used to compute this score at
application time. In our mixed configuration $M{=}9$: the four LLMs in
zero-shot and few-shot settings (eight prompted variants) plus the
classical TF-IDF + LR baseline. Because this pool mixes the eight
LLM variants with the stronger classical baseline, an LLM's
agreement partly reflects concordance with a more accurate
supervised predictor, not peer consensus alone. We therefore also
report an LLM-only $M{=}8$ control. \emph{Operational trade-off:}
agreement requires
running all $M$ predictors at inference. Because the pool is
heterogeneous, its dollar cost and latency depend on the endpoints
and are not necessarily $M{\times}$ those of the primary alone.
Isotonic incurs one calibration-time pass per prompted predictor and a
constant-time lookup at inference.

\textit{Isotonic regression on a held-out calibration split.}
LLM-emitted \texttt{confidence} fields are bounded in $[0,1]$ but
are not logits, so parametric Platt or temperature scaling do not
apply cleanly. We instead fit one isotonic regression per prompted
LLM condition
on the calibration split: pairs
$\{(c_i, \mathbf{1}[\hat{y}_i = y_i])\}$ of raw confidence and
ground-truth correctness, mapped via
\texttt{sklearn.isotonic.IsotonicRegression}
\cite{pedregosa2011scikit} (monotone non-decreasing, output
clipped to $[0,1]$) into a recalibration
$c \mapsto \hat{\Pr}(\text{correct})$. For each LLM predictor we
fit on a stratified 500-ticket subsample of the
calibration split (4{,}000 calibration queries across the 8
prompted variants). \textbf{Calibration and test splits are
disjoint}: the isotonic mapping is learned only from calibration
pairs and applied unchanged to test confidences. ECE is computed
with 10 equal-width bins on test predictions.

\textit{Discrimination metrics.} ECE measures probability-level
calibration but not whether confidence ranks correct predictions
above errors. We therefore report correctness AUROC (ties receive
half credit) and area under the risk--coverage curve (AURC; lower is
better). Since LLM scores contain many ties, our AURC adds each
equal-score group as a block and integrates the right-continuous risk
step, avoiding arbitrary ordering within ties.

\textit{Nontrivial rejection comparators.} For TF-IDF + LR we also
use the top-two probability margin $p_{(1)}-p_{(2)}$ and normalized
inverse entropy $1-H(p)/\log K$. Each score's threshold is selected
on calibration over a 0.01-spaced grid in $[0,1]$ and then evaluated
once on test.

\subsection{Abstention policies}
For the primary non-LLM analysis we compare
\texttt{always\_route} (route every ticket) with
\texttt{threshold@}$\tau$ for $\tau \in \{0.5, 0.6, 0.7, 0.8, 0.9\}$
(route if $\text{conf} \geq \tau$, defer otherwise). Where stated, we
also use the trivial \texttt{always\_defer} reference policy, which
defers every ticket at cost $c_{\mathrm{triage}}$ by construction. It
appears in the LLM break-even comparison and the post-hoc cost grid;
the primary non-LLM analysis uses \texttt{always\_route} as its
baseline. Cross-predictor agreement enters our experiments only
as a confidence estimator (above), not as a deferral policy.

\textit{Threshold selection.} Non-LLM operating points are selected
on the calibration split under the configured cost model. The best
threshold over the stated grid,
$\tau^* = \arg\min_\tau \mathbb{E}_{\text{calib}}[c \mid \tau]$,
is then evaluated once on the held-out test split. For the
classical TF-IDF + LR baseline this procedure selects
$\tau^* = 0.60$ (calibration
costs at $\tau \in \{0.5, 0.6, 0.7, 0.8, 0.9\}$ are
$\{0.493, 0.461, 0.477, 0.537, 0.654\}$). Calibration and test
splits are disjoint by construction.

For each post-hoc cost-sensitivity setting, we select on calibration
among \texttt{always\_route}, \texttt{always\_defer}, and thresholds
from 0.01 to 0.99 in steps of 0.01, then evaluate the frozen choice
once on test. For exact calibration-cost ties, the deterministic order
is \texttt{always\_route}, then \texttt{always\_defer}, then the
smallest threshold among tied threshold policies.

For LLMs we do not select a threshold on test. If an isotonic score
is interpreted as $p=\Pr(\mathrm{correct})$, auto-routing is
cost-preferred when
\begin{equation}
p c_{\mathrm{ok}}+(1-p)c_{\mathrm{wrong}}
\leq c_{\mathrm{triage}},
\label{eq:breakeven}
\end{equation}
which gives the fixed break-even threshold $p\geq0.80$ under
$(0,1,5)$. Raw verbalized scores are not treated as probabilities
for policy optimization.

\subsection{Modular evaluation pipeline}
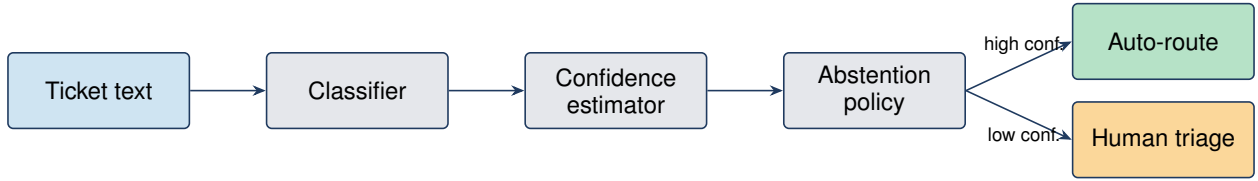
\begin{figure*}[t]
  \centering
  \begin{tikzpicture}[
      font=\small\sffamily,
      node distance=6mm and 10mm,
      box/.style={
        rounded corners=2pt, draw=rgStroke, line width=0.6pt,
        minimum height=10mm, minimum width=24mm, align=center,
        inner sep=3pt,
      },
      input/.style={box, fill=rgInput},
      stage/.style={box, fill=rgTransform},
      auto/.style={box, fill=rgAuto},
      defer/.style={box, fill=rgDefer},
      arr/.style={-{Stealth[length=5pt,width=4pt]}, line width=0.6pt, draw=rgStroke},
    ]
    \node[input] (text) {Ticket text};
    \node[stage, right=of text] (clf) {Classifier};
    \node[stage, right=of clf] (conf) {Confidence\\estimator};
    \node[stage, right=of conf] (pol) {Abstention\\policy};
    \node[auto,  right=14mm of pol, yshift=6.5mm] (auto) {Auto-route};
    \node[defer, right=14mm of pol, yshift=-6.5mm] (defer) {Human triage};

    \draw[arr] (text) -- (clf);
    \draw[arr] (clf)  -- (conf);
    \draw[arr] (conf) -- (pol);
    \draw[arr] (pol.east) -- node[above, font=\scriptsize\sffamily, pos=0.55] {high conf.} (auto.west);
    \draw[arr] (pol.east) -- node[below, font=\scriptsize\sffamily, pos=0.55] {low conf.}  (defer.west);
  \end{tikzpicture}
  \caption{Evaluation pipeline from ticket text to auto-routing or human
  triage, with prediction, confidence estimation, and abstention separated.}
  \label{fig:pipeline}
\end{figure*}

Predictors, confidence estimators, abstention policies, and the
cost model are separated by interfaces. Registry-based predictor
addition does not require changing the evaluator; other extension
types require explicit builder or policy-set wiring in the current code.
Fig.~\ref{fig:pipeline} shows the flow.

\section{Experimental Setup}
\label{sec:setup}

\subsection{Dataset and splits}
\begin{figure}[t]
  \centerline{\includegraphics[width=0.95\columnwidth]{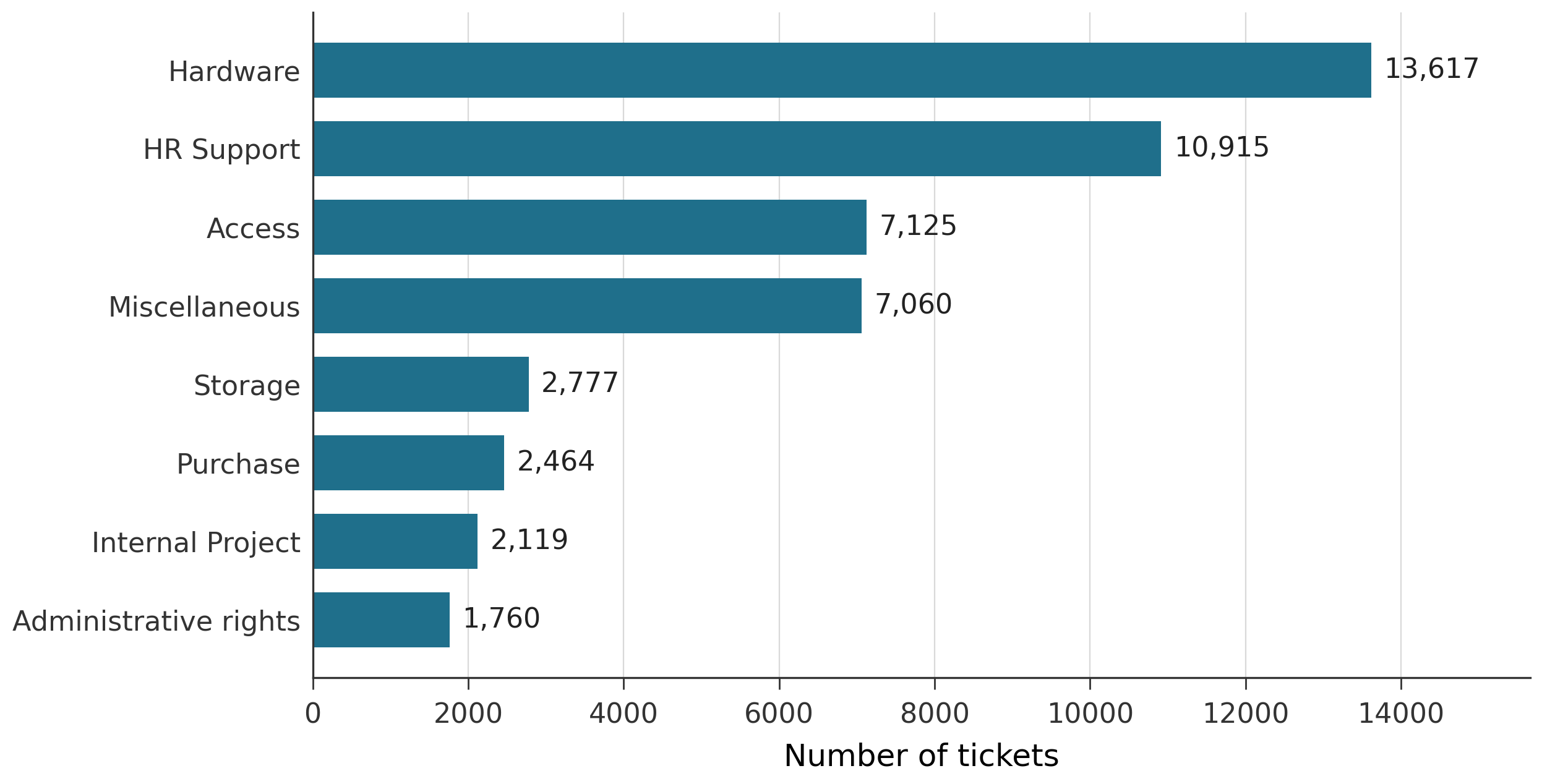}}
  \caption{Class distribution of the 47{,}837-ticket corpus across
  the eight categories.}
  \label{fig:classdist}
\end{figure}

We use the public IT Service Ticket Classification Dataset
\cite{itservicetickets}: 47{,}837 tickets, 8 categories
(\emph{Access}, \emph{Administrative rights}, \emph{HR Support},
\emph{Hardware}, \emph{Internal Project}, \emph{Miscellaneous},
\emph{Purchase}, \emph{Storage}), median ticket length 175
characters, maximum/minimum class size 13{,}617 / 1{,}760
($7.74\times$ imbalance; Fig.~\ref{fig:classdist}). We use a stratified 70/10/20
train/calibration/test split (Table~\ref{tab:dataset}). Few-shot
exemplars are drawn only from the training split. The dataset is
publicly available on Kaggle under its posted terms. We use that
preprocessed public release. The complete dataset and per-ticket text
arrays are not redistributed in the camera-ready artifacts; only three
brief illustrative excerpts from the public dataset are quoted in
Section~\ref{sec:error-analysis}. Each dataset row pairs a short internal
IT request with a category label.

\begin{table}[t]
\caption{Dataset summary (public IT Service Ticket Classification Dataset).}
\label{tab:dataset}
\centering
\begin{tabular}{lr}
\toprule
Property & Value \\
\midrule
\# tickets        & 47{,}837 \\
\# classes        & 8 \\
Train / Calib. / Test & 33{,}486 / 4{,}784 / 9{,}567 \\
Median ticket length (chars) & 175 \\
Class imbalance (max/min) & $7.74\times$ \\
\bottomrule
\end{tabular}
\end{table}

\subsection{Test subsample for LLM predictors}
Per-call API budget makes evaluating all four LLMs on the full
9{,}567-ticket test set impractical. The 8 LLM configurations
alone issue $8{,}000$ test queries on the subsample described
below, and scaling to the full test set would multiply that by an
order of magnitude. LLMs are therefore evaluated on a stratified
1{,}000-ticket subsample (seed 42). To assess whether that subset
preserves the non-LLM baseline ordering, we re-evaluate every
non-LLM baseline on the same
1{,}000-ticket subsample (Table~\ref{tab:subsample}). Subsample
and full-test numbers differ by at most 1.6 percentage points for
every baseline. The strongest non-LLM baseline (DistilBERT)
reaches 0.879 accuracy on the subsample, versus 0.477 for the
best LLM on the same tickets, confirming the same ordering on
identical examples.

\begin{table}[t]
\caption{Non-LLM baselines on the LLM test subset ($n=1{,}000$).}
\label{tab:subsample}
\centering
\footnotesize
\setlength{\tabcolsep}{4pt}
\begin{tabular}{@{}lccc@{}}
\toprule
Baseline (on 1{,}000-ticket subsample) & Acc. & Macro-F1 & ECE-raw \\
\midrule
TF-IDF + LR            & 0.847 & 0.842 & 0.111 \\
TF-IDF + Linear SVM    & 0.867 & 0.869 & 0.056 \\
TF-IDF + Random Forest & 0.804 & 0.791 & 0.172 \\
DistilBERT (fine-tuned) & \textbf{0.879} & \textbf{0.883} & \textbf{0.043} \\
\bottomrule
\end{tabular}
\end{table}

\subsection{Reproducibility}
Runs save configuration, seeds, dataset and split hashes, package
versions, and per-ticket outputs. Intervals use 1{,}000 paired
resamples with seed 1337; reported percentage reductions recompute
the ratio inside each resample. The public project repository is
\url{https://github.com/sachinkg12/RouteGuard}; the frozen
camera-ready claim manifest, automated recomputation receipt from a
separate verification script,
tables, text-free per-ticket vectors, and code snapshot are archived
as a versioned series at \href{https://doi.org/10.5281/zenodo.22609128}%
{doi:10.5281/zenodo.22609128}.
The supplement pins the verified x86-64/Python 3.14 environment
because estimator refits can vary slightly across architectures;
the frozen prediction vectors remain the audit record.

\section{Results}
\label{sec:results}

\subsection{Base-classifier predictive performance}
\label{sec:results-classification}

We compare four non-LLM baselines against eight prompted hosted LLM
configurations (Table~\ref{tab:results}). Among non-LLM baselines
the fine-tuned encoder is strongest. DistilBERT reaches accuracy
\textbf{0.877} and macro-F1 \textbf{0.874} on the full
9{,}567-ticket test set after two epochs of training on a single
GPU with batch size 32 and maximum sequence length 256. The next
non-LLM baseline is a sigmoid-calibrated linear SVM over TF-IDF
features (accuracy \textbf{0.863}, macro-F1 \textbf{0.864}),
followed by the TF-IDF + logistic regression baseline
(0.847 / 0.842) and a TF-IDF random forest
(0.818 / 0.806).

The best prompted LLM is Claude Haiku 4.5 with $k{=}3$ few-shot
prompting at $0.477 / 0.459$. GPT-4o-mini few-shot follows
(0.459 / 0.450), then Qwen2.5-7B-Instruct few-shot (0.447 / 0.454)
and Llama-3-8B-Instruct-Lite zero-shot (0.378 / 0.284). On the
common 1{,}000-ticket subset, DistilBERT reaches 0.879 and Haiku
few-shot 0.477, so the measured gap is not caused by different test
examples. It is not, however, a pure capability comparison: the
trained models learn the taxonomy from 33{,}486 labeled tickets,
whereas the prompted LLMs receive label names and, at most, 24
examples without category definitions. Ambiguous labels such as
\emph{Miscellaneous}, \emph{Internal Project}, and
\emph{Administrative rights} make that distinction material. We
therefore scope the result to these training and prompt-information
regimes.

Few-shot prompting helps three of the four LLMs but \emph{hurts}
Llama-3-8B-Instruct-Lite by 12.3 percentage points. Inspection of the
raw outputs shows that the model often emits a conversational preamble (\emph{``I'll
classify the ticket and return the JSON.''}) that uses up the
output budget before any JSON is emitted, which produces a
15.1\,\% parse-error rate.\footnote{This low few-shot result is
sensitive to the joint prompt/output-budget configuration: on the
ablation subsample, switching from \emph{v1}/64 to the stricter
\emph{v2}/128 configuration eliminates the
parse failures and raises Llama few-shot accuracy from 0.245 to
0.320 (Table~\ref{tab:ablation}).} We dig into this in
\S\ref{sec:results-ablation}.

\begin{table*}[t]
\caption{Classification and confidence results. Non-LLM: full test;
LLM: common 1{,}000-ticket subset. $r/i$: raw/isotonic (fit on
calibration); ECE and tie-group AURC $\downarrow$, AUROC $\uparrow$.
Bold marks the best value per metric within each evaluation block.}
\label{tab:results}
\centering
\scriptsize
\setlength{\tabcolsep}{2.5pt}
\begin{tabular}{llcccccccc}
\toprule
Model & Setting & Acc. & Macro-F1 & ECE$_r$ & ECE$_i$ & AUROC$_r$ & AUROC$_i$ & AURC$_r$ & AURC$_i$ \\
\midrule
TF-IDF + LR              & classical & 0.847 & 0.842 & 0.107 & --- & --- & --- & --- & --- \\
TF-IDF + Linear SVM      & classical & 0.863 & 0.864 & 0.046 & --- & --- & --- & --- & --- \\
TF-IDF + Random Forest   & classical & 0.818 & 0.806 & 0.182 & --- & --- & --- & --- & --- \\
DistilBERT (fine-tuned)  & classical & \textbf{0.877} & \textbf{0.874} & \textbf{0.041} & --- & --- & --- & --- & --- \\
\midrule
Claude Haiku 4.5      & few-shot  & \textbf{0.477} & \textbf{0.459} & 0.386 & 0.035 & 0.664 & 0.659 & \textbf{0.417} & \textbf{0.430} \\
GPT-4o-mini           & few-shot  & 0.459 & 0.450 & 0.541 & 0.021 & 0.500 & 0.500 & 0.541 & 0.541 \\
Qwen2.5-7B-Instruct   & few-shot  & 0.447 & 0.454 & 0.524 & 0.049 & 0.509 & 0.509 & 0.548 & 0.549 \\
Llama-3-8B-Inst.-Lite & zero-shot & 0.378 & 0.284 & 0.428 & 0.025 & 0.513 & 0.513 & 0.616 & 0.616 \\
GPT-4o-mini           & zero-shot & 0.376 & 0.372 & 0.456 & 0.030 & 0.655 & 0.655 & 0.537 & 0.537 \\
Claude Haiku 4.5      & zero-shot & 0.360 & 0.364 & \textbf{0.332} & 0.047 & \textbf{0.671} & \textbf{0.678} & 0.526 & 0.541 \\
Qwen2.5-7B-Instruct   & zero-shot & 0.355 & 0.363 & 0.468 & 0.032 & 0.555 & 0.555 & 0.619 & 0.618 \\
Llama-3-8B-Inst.-Lite & few-shot  & 0.255 & 0.266 & 0.594 & \textbf{0.014} & 0.601 & 0.601 & 0.706 & 0.706 \\
\bottomrule
\end{tabular}
\end{table*}

\subsection{Is the confidence signal useful?}
\label{sec:results-calibration}

Every prompted LLM is overconfident: raw ECE spans
$[0.332,0.594]$, and GPT-4o-mini few-shot assigns confidence 1.000
to every ticket despite 0.459 accuracy. Isotonic regression fitted
on the disjoint calibration subset reduces LLM ECE to
$[0.014,0.049]$, an 86--98\,\% relative reduction.

That ECE result must not be read as improved error ranking.
Isotonic is monotone and changes correctness AUROC by at most 0.007
in these runs; AURC also does not improve consistently
(Table~\ref{tab:results}). The limiting example is GPT-4o-mini
few-shot: raw confidence is the constant 1.000, isotonic maps it to
the constant 0.480, and both correctness AUROC and AURC remain
0.500 and 0.541. Isotonic corrects the probability level by mapping
a non-discriminative score to the base accuracy; it does not create
information that was absent from the score.

Agreement can carry ranking signal, but the voter pool matters.
Table~\ref{tab:agreement-control} separates the mixed $M{=}9$ pool
from the LLM-only $M{=}8$ control. The $M{=}8$ correctness AUROC
spans 0.590--0.773, descriptive evidence that peer-only agreement
contains ranking signal on these data. Its ECE also improves over
raw confidence in all eight conditions, but ECE alone would not
establish discrimination. The mixed pool is better in every listed
ECE and AUROC cell, but it includes the supervised TF-IDF + LR vote.
Thus $M{=}9$ uses no held-out calibration labels when a score is
computed, yet it is not free of supervised information. It also
requires all predictors at inference.

\begin{table*}[t]
\caption{Mixed and LLM-only agreement controls on the common
1{,}000-ticket test subset. The mixed $M{=}9$ pool adds supervised
TF-IDF + LR to the eight prompted conditions.}
\label{tab:agreement-control}
\centering
\footnotesize
\setlength{\tabcolsep}{4pt}
\begin{tabular}{llcccc}
\toprule
Model & Setting & ECE$_{M9}$ & ECE$_{M8}$ & AUROC$_{M9}$ & AUROC$_{M8}$ \\
\midrule
Claude Haiku 4.5      & few-shot  & 0.148 & 0.210 & 0.692 & 0.590 \\
GPT-4o-mini           & few-shot  & 0.205 & 0.239 & 0.707 & 0.593 \\
Qwen2.5-7B-Instruct   & few-shot  & 0.196 & 0.238 & 0.708 & 0.603 \\
Llama-3-8B-Inst.-Lite & zero-shot & 0.193 & 0.209 & 0.750 & 0.658 \\
GPT-4o-mini           & zero-shot & 0.271 & 0.305 & 0.800 & 0.694 \\
Claude Haiku 4.5      & zero-shot & 0.251 & 0.283 & 0.814 & 0.721 \\
Qwen2.5-7B-Instruct   & zero-shot & 0.264 & 0.299 & 0.814 & 0.718 \\
Llama-3-8B-Inst.-Lite & few-shot  & 0.212 & 0.239 & 0.840 & 0.773 \\
\bottomrule
\end{tabular}
\end{table*}

\subsection{Which routing policy is cost-preferred?}
\label{sec:results-cost}

\paragraph{Headline result on the classical baseline}
We illustrate the cost framework on the TF-IDF + LR baseline
because it is the simplest and most reproducible substrate.
The same machinery applies to any predictor in
Table~\ref{tab:results}: stronger non-LLM baselines (DistilBERT,
linear SVM) defer less aggressively but follow the same
qualitative shape, since they enter the cost calculation through
a better-calibrated, more accurate confidence signal.

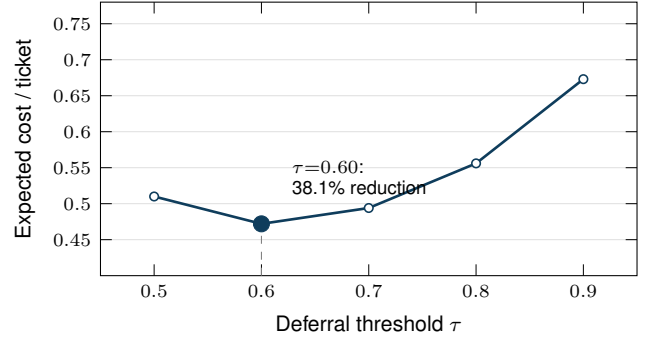
\begin{figure}[t]
  \centering
  \begin{tikzpicture}
    \begin{axis}[
        width=0.98\columnwidth, height=5.2cm,
        xlabel={Deferral threshold $\tau$},
        ylabel={Expected cost / ticket},
        xmin=0.45, xmax=0.95,
        ymin=0.40, ymax=0.78,
        xtick={0.5,0.6,0.7,0.8,0.9},
        ytick={0.45,0.50,0.55,0.60,0.65,0.70,0.75},
        ymajorgrids, grid style={gray!25, line width=0.3pt},
        tick style={black, line width=0.3pt},
        axis line style={black, line width=0.4pt},
        label style={font=\footnotesize\sffamily},
        tick label style={font=\scriptsize\sffamily},
        clip=false,
      ]
      \addplot+[
        mark=*, mark size=1.6pt,
        line width=1.0pt, color=rgLine,
        mark options={fill=white, draw=rgLine, line width=0.6pt},
      ] coordinates {
        (0.50, 0.510)
        (0.60, 0.472)
        (0.70, 0.494)
        (0.80, 0.556)
        (0.90, 0.673)
      };
      \addplot[only marks, mark=*, mark size=3pt,
        mark options={fill=rgLine, draw=rgLine}]
        coordinates {(0.60, 0.472)};
      \draw[dashed, gray, line width=0.4pt]
        (axis cs:0.60,0.40) -- (axis cs:0.60,0.472);
      \node[font=\scriptsize\sffamily, anchor=south west, align=left]
        at (axis cs:0.62, 0.50)
        {$\tau{=}0.60$:\\38.1\% reduction};
    \end{axis}
  \end{tikzpicture}
  \caption{TF-IDF + LR test cost over the predeclared threshold
  grid under $(0,1,5)$. The calibration-selected $\tau=0.60$
  yields a 38.1\,\% reduction (95\,\% CI
  $[35.7\,\%, 40.4\,\%]$).}
  \label{fig:costthresh}
\end{figure}

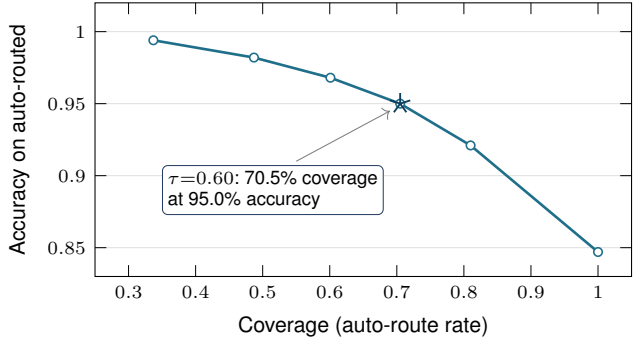
\begin{figure}[t]
  \centering
  \begin{tikzpicture}
    \begin{axis}[
        width=0.98\columnwidth, height=5.2cm,
        xlabel={Coverage (auto-route rate)},
        ylabel={Accuracy on auto-routed},
        xmin=0.25, xmax=1.05,
        ymin=0.83, ymax=1.02,
        xtick={0.3,0.4,0.5,0.6,0.7,0.8,0.9,1.0},
        ytick={0.85,0.90,0.95,1.00},
        ymajorgrids, grid style={gray!25, line width=0.3pt},
        tick style={black, line width=0.3pt},
        axis line style={black, line width=0.4pt},
        label style={font=\footnotesize\sffamily},
        tick label style={font=\scriptsize\sffamily},
        clip=false,
      ]
      \addplot+[
        mark=*, mark size=1.6pt,
        line width=1.0pt, color=rgTeal,
        mark options={fill=white, draw=rgTeal, line width=0.6pt},
      ] coordinates {
        (1.00, 0.847)
        (0.81, 0.921)
        (0.705, 0.950)
        (0.601, 0.968)
        (0.487, 0.982)
        (0.337, 0.994)
      };
      \addplot[only marks, mark=star, mark size=4pt,
        mark options={draw=rgLine, line width=0.7pt, fill=rgLine}]
        coordinates {(0.705, 0.950)};
      \node[font=\scriptsize\sffamily, anchor=west, align=left,
            draw=rgStroke, rounded corners=2pt, fill=white,
            inner sep=2pt, line width=0.3pt]
        at (axis cs:0.35, 0.89)
        {$\tau{=}0.60$: 70.5\% coverage\\at 95.0\% accuracy};
      \draw[->, gray, line width=0.4pt]
        (axis cs:0.55, 0.91) -- (axis cs:0.69, 0.945);
    \end{axis}
  \end{tikzpicture}
  \caption{TF-IDF + LR coverage--accuracy curve. The marked
  calibration-selected point routes 70.5\,\% of test tickets at
  95.0\,\% accuracy.}
  \label{fig:covacc}
\end{figure}

At the illustrative cost setting $(0, 1, 5)$, the calibration-selected
threshold $\tau^* = 0.60$ reduces expected cost per ticket from
$0.763$ to $0.472$ on the held-out test split, a relative reduction
of \textbf{38.1\,\%} (95\,\% paired-bootstrap CI
\textbf{[35.7\,\%, 40.4\,\%]}). The ratio is recomputed inside
each paired bootstrap resample; the corresponding absolute cost
difference is $-0.291$ (95\,\% CI $[-0.318,-0.263]$) per ticket.
The wrong-route rate
drops from $15.3\,\%$ to $3.5\,\%$ while the system still
auto-routes $70.5\,\%$ of tickets at $95.0\,\%$ on-routed accuracy
(Fig.~\ref{fig:covacc}). The evaluated cost points form a U-shape
with a minimum at $\tau^*$ (Table~\ref{tab:cost},
Fig.~\ref{fig:costthresh}), and the
bootstrap interval excludes zero by a wide margin
(Table~\ref{tab:cost-baselines}).

\begin{table}[t]
\caption{TF-IDF + LR test cost under $(0,1,5)$. Bold marks the
calibration-selected operating point; reductions are relative to
\texttt{always\_route}.}
\label{tab:cost}
\centering
\footnotesize
\setlength{\tabcolsep}{3.5pt}
\begin{tabular}{@{}lcccc@{}}
\toprule
Policy & $\mathbb{E}[c]$ & Reduction\,\% & Cov.\,\% & Acc.\,\% \\
\midrule
\texttt{always\_route}     & 0.763 & ---      & 100.0 & 84.7 \\
\texttt{threshold@0.50}    & 0.510 & 33.1 & 81.0 & 92.1 \\
\textbf{\texttt{threshold@0.60}} & \textbf{0.472} & \textbf{38.1} & \textbf{70.5} & \textbf{95.0} \\
\texttt{threshold@0.70}    & 0.494 & 35.2 & 60.1 & 96.8 \\
\texttt{threshold@0.80}    & 0.556 & 27.2 & 48.7 & 98.2 \\
\texttt{threshold@0.90}    & 0.673 & 11.8 & 33.7 & 99.4 \\
\bottomrule
\end{tabular}
\end{table}

\paragraph{Cost analysis is consistent across non-LLM baselines}
We re-run the same procedure (per-model calibration-selected
$\tau^*$, evaluated on the held-out test split) for the SVM,
random-forest, and DistilBERT baselines. Table~\ref{tab:cost-baselines}
reports the results. The relative cost reduction at
each model's $\tau^*$ clusters in a narrow band of 35.5 to 38.1\,\%,
with overlapping bootstrap intervals, even though the always-route
baseline cost varies substantially (from 0.615 for DistilBERT to
0.910 for the random forest) and the selected $\tau^*$ varies from
0.60 to 0.90: LR and random forest select 0.60, SVM selects 0.70,
and DistilBERT selects 0.90. Because these thresholds also depend on
each model's score scale and calibration, they do not establish a
general relationship between accuracy and threshold. DistilBERT's
choice is the edge value on the predeclared grid. A post-hoc
0.01-spaced sensitivity sweep moves the
calibration choice to 0.94 and gives test cost 0.401 (34.7\,\%
reduction), close to the primary 0.396 (35.5\,\%) result. Thus grid
density changes the operating point but not the qualitative
conclusion; the dense sweep does not replace the predeclared analysis.

\begin{table}[t]
\caption{Non-LLM test costs at calibration-selected thresholds under
$(0,1,5)$. $\mathbb{E}[c]_{\mathrm{a}}$: always-route cost; Wr.:
post-threshold wrong-route rate; ratio-bootstrap 95\,\% CIs.}
\label{tab:cost-baselines}
\centering
\scriptsize
\setlength{\tabcolsep}{2.5pt}
\begin{tabular}{@{}lcccccc@{}}
\toprule
Baseline & $\tau^*$ & $\mathbb{E}[c]_{\text{a}}$ & $\mathbb{E}[c]_{\tau^*}$ & Reduction\,\% & 95\,\% CI & Wr.\,\% \\
\midrule
TF-IDF + LR            & 0.60 & 0.763 & 0.472 & 38.1 & $[35.7,40.4]$ & 3.5 \\
TF-IDF + Linear SVM    & 0.70 & 0.687 & 0.434 & 36.7 & $[34.3,39.1]$ & 3.7 \\
TF-IDF + Random Forest & 0.60 & 0.910 & 0.571 & 37.3 & $[35.0,39.5]$ & 2.1 \\
DistilBERT (fine-tuned) & 0.90 & 0.615 & 0.396 & 35.5 & $[32.9,38.2]$ & 3.3 \\
\bottomrule
\end{tabular}
\end{table}

\paragraph{Cost-parameter sensitivity}
Table~\ref{tab:cost-sensitivity} varies both uncertain costs while
selecting one policy per setting on calibration. A threshold policy is selected
in 15 of 16 settings. On held-out test data, the frozen choice costs
less than both trivial policies in 14 settings, and all 14 paired
ratio-bootstrap 95\,\% intervals exclude zero. At the equal-cost
$(2,2)$ boundary, calibration selects \texttt{always\_route}. At
$(1.5,2)$, the calibration-selected threshold is 0.2\,\% costlier
than the lower-cost trivial test comparator, illustrating finite-split
selection variation rather than a universal gain. As the wrong-route
cost grows relative to triage, the selected threshold generally rises.
The $(1,5)$ cell uses the post-hoc dense grid and therefore matches
the 36.9\,\% top-probability result in Table~\ref{tab:rejection}; it
does not replace the primary five-threshold 38.1\,\% result.

\begin{table*}[t]
\caption{Post-hoc TF-IDF + LR cost-parameter sensitivity. For each
setting, one policy is selected on calibration over \texttt{always\_route},
\texttt{always\_defer}, and thresholds 0.01--0.99, then evaluated once
on test. Cells show policy / test $\mathbb{E}[c]$ / reduction relative
to the lower-cost trivial test comparator. Negative reduction denotes
higher cost. \texttt{t@x} abbreviates \texttt{threshold@x}; $\dagger$
marks the equal-cost boundary.}
\label{tab:cost-sensitivity}
\centering
\scriptsize
\setlength{\tabcolsep}{4pt}
\begin{tabular}{@{}c*{4}{c}@{}}
\toprule
$c_{\mathrm{triage}}\backslash c_{\mathrm{wrong}}$ & 2 & 5 & 10 & 20 \\
\midrule
0.5 & \texttt{t@0.55} / 0.220 / 27.9\% & \texttt{t@0.71} / 0.293 / 41.4\% & \texttt{t@0.79} / 0.342 / 31.6\% & \texttt{t@0.88} / 0.374 / 25.2\% \\
1.0 & \texttt{t@0.39} / 0.298 / 2.4\%  & \texttt{t@0.65} / 0.482 / 36.9\% & \texttt{t@0.71} / 0.586 / 41.4\% & \texttt{t@0.79} / 0.684 / 31.6\% \\
1.5 & \texttt{t@0.22} / 0.306 / $-0.2$\% & \texttt{t@0.53} / 0.606 / 20.6\% & \texttt{t@0.65} / 0.789 / 47.4\% & \texttt{t@0.79} / 0.934 / 37.7\% \\
2.0 & \texttt{always\_route}$^{\dagger}$ / 0.305 / 0.0\% & \texttt{t@0.42} / 0.698 / 8.5\% & \texttt{t@0.65} / 0.963 / 36.9\% & \texttt{t@0.71} / 1.173 / 41.4\% \\
\bottomrule
\end{tabular}
\end{table*}

\paragraph{Rejection-score comparators}
Table~\ref{tab:rejection} compares rejection scores. Its post-hoc
0.01-grid top-probability threshold (0.65) yields 36.9\,\% on test,
versus 38.1\,\% at 0.60 on the primary predeclared grid; this
calibration-to-test variation does not replace the primary result.
Margin is slightly better and entropy worse, but each CI overlaps
the top-probability CI. Because the intervals are not paired between
rules, their ordering is descriptive. All three reduce cost.

\begin{table}[t]
\caption{TF-IDF + LR rejection scores under $(0,1,5)$. Thresholds are
selected on calibration over a 0.01 grid and evaluated on test. Cost
reductions and paired ratio-bootstrap 95\,\% CIs are relative to
always-route.}
\label{tab:rejection}
\centering
\scriptsize
\setlength{\tabcolsep}{2.5pt}
\begin{tabular}{@{}lccccc@{}}
\toprule
Score & $\tau^*$ & $\mathbb{E}[c]$ & Reduction\,\% & 95\,\% CI & Cov.\,\% \\
\midrule
Top probability & 0.65 & 0.482 & 36.9 & $[34.2,39.3]$ & 65.2 \\
Top-two margin  & 0.43 & 0.467 & 38.8 & $[36.5,41.1]$ & 68.5 \\
Inverse entropy & 0.44 & 0.507 & 33.6 & $[30.8,36.2]$ & 68.3 \\
\bottomrule
\end{tabular}
\end{table}

\paragraph{LLM break-even policies}
We evaluate the isotonic scores at the break-even probability 0.80
derived in \eqref{eq:breakeven}; it is fixed by the cost
model, not selected from test outcomes. Seven of eight prompted
conditions route 0/1{,}000 tickets and therefore equal the
always-defer cost of 1.0. Claude Haiku few-shot routes 19/1{,}000
(1.9\,\% coverage); all 19 are correct, giving cost 0.981 and a
1.9\,\% reduction versus always-defer (paired ratio-bootstrap
95\,\% CI [1.1\,\%, 2.8\,\%]). Because all 19 routed tickets are
correct, this bootstrap interval reflects observed coverage variation
and cannot generate an unseen routed error. The exact binomial 95\,\%
CI for routed accuracy is [82.4\,\%, 100\,\%], with its lower bound
only 2.4 percentage points above the 80\,\% break-even. This is a low-coverage observation,
not evidence of a broadly useful LLM routing policy. Raw verbalized
confidence is not used for this decision because its probability
interpretation is contradicted by the ECE results.

The outcome is conditional on $(0,1,5)$. Different local costs move
the analytic break-even point, and local prevalence or drift can
change calibration. We therefore treat defer-all as the go/no-go
outcome for seven evaluated LLM conditions, not as a universal claim
about these model families.

\subsection{Prompt/budget sensitivity}
\label{sec:results-ablation}

We probe the brittleness seen in
\S\ref{sec:results-classification} by comparing two joint
prompt/output-budget configurations on Llama-3-8B-Instruct-Lite,
evaluated on a stratified 200-ticket subsample
(Table~\ref{tab:ablation}). The two configurations change the
prompt template and the token budget \emph{together}: \emph{v1/64}
(the verbose main-run prompt at a 64-token cap) versus
\emph{v2/128} (a stricter ``JSON-only, no preamble'' prompt at a
128-token cap), so the experiment measures their combined effect
rather than isolating prompt format. Switching from v1/64 to
v2/128 removes few-shot parse errors (18.5\,\% to 0\,\%) and
improves few-shot accuracy by $+7.5$ percentage points. The same change
\emph{hurts} zero-shot accuracy by $-9.5$ percentage points, as the model instead
falls back to the catch-all \emph{Miscellaneous} label on 77.5\,\%
of inputs. No single configuration wins across both settings for
this model. This setting-dependent sensitivity is another
operational reason to place a deferral guardrail around this
evaluated Llama configuration rather than auto-route unguarded.

\begin{table}[t]
\caption{Joint prompt/budget sensitivity for Llama-3-8B on a
200-ticket subset. Template and token budget change together. Bold
marks the higher accuracy within each zero-/few-shot pair.}
\label{tab:ablation}
\centering
\footnotesize
\setlength{\tabcolsep}{3.5pt}
\begin{tabular}{@{}lccccc@{}}
\toprule
Setting & Prompt & Max tok. & Acc. & Macro-F1 & Parse err. \\
\midrule
zero-shot & v1 & 64  & \textbf{0.345} & 0.245 & 0.000 \\
zero-shot & v2 & 128 & 0.250 & 0.197 & 0.000 \\
few-shot  & v1 & 64  & 0.245 & 0.271 & 0.185 \\
few-shot  & v2 & 128 & \textbf{0.320} & 0.289 & 0.000 \\
\bottomrule
\end{tabular}
\end{table}

\subsection{Alternate-provider/precision check}
\label{sec:results-fullprec}
The primary open-model endpoints were documented by Together as FP8
and INT4; response metadata did not record precision
(\S\ref{sec:method}). We re-ran the corresponding model families
through DeepInfra, documented by that provider as BF16, on the same
1{,}000-ticket subsample (
\texttt{Qwen/Qwen2.5-7B-Instruct} and
\texttt{meta-llama/Meta-Llama-3-8B-Instruct}, queried 2026-07-17,
temperature 0, \texttt{max\_tokens}=64, same \emph{v1} template).
Because provider and provider-documented precision changed together,
this is not a controlled quantization ablation. Qwen barely moves (zero-shot
$0.355 \to 0.369$, few-shot $0.447 \to 0.459$; parse errors
$0\,\%$ throughout). Llama moves more and unevenly: few-shot
recovers substantially ($0.255 \to 0.409$, with the parse-error
rate falling $15.1 \to 3.3\,\%$) despite the unchanged \emph{v1}/64
prompt/output-budget configuration, while
zero-shot does not improve ($0.378 \to 0.319$; parse errors
$0 \to 0.8\,\%$). The best alternate-provider open-model result (0.459)
still trails the best prompted LLM (Claude Haiku few-shot, 0.477)
and remains far below the classical baselines, so our conclusions
are unchanged.

\subsection{Error analysis}
\label{sec:error-analysis}
\begin{figure}[t]
  \centerline{\includegraphics[width=0.95\columnwidth]{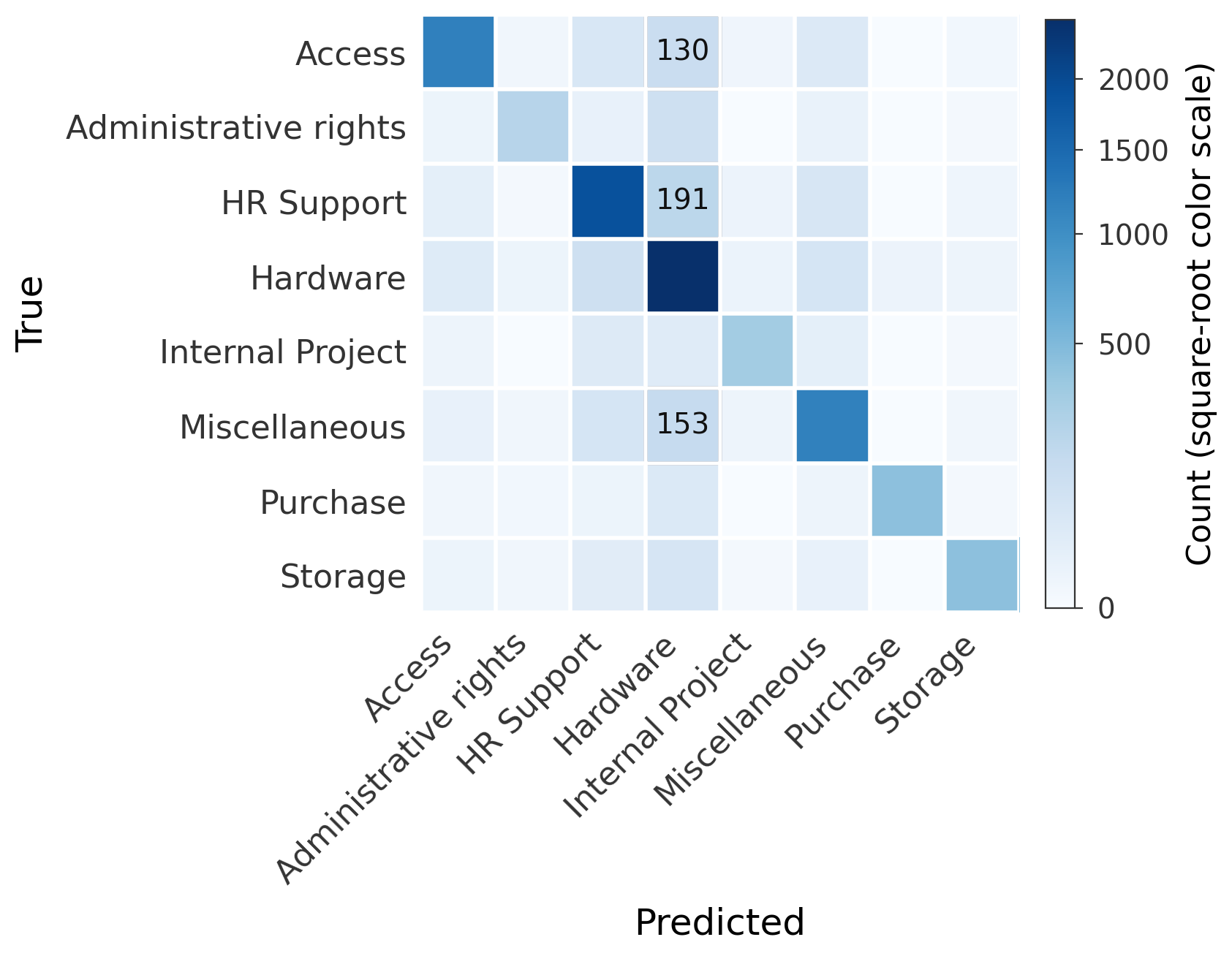}}
  \caption{TF-IDF + LR confusion matrix without abstention on the
  full test set ($n=9{,}567$). Color uses a square-root count scale;
  the three largest off-diagonal cells are labeled.}
  \label{fig:confusion}
\end{figure}

The most frequent confusion pair for the TF-IDF + LR baseline is
\emph{HR Support $\to$ Hardware} (191 tickets), followed by
\emph{Miscellaneous $\to$ Hardware} (153) and
\emph{Access $\to$ Hardware} (130). These are typically short
tickets with generic IT vocabulary that lacks the lexical anchors
this model relies on. Fig.~\ref{fig:confusion} shows its confusion
matrix without abstention.

LLM errors show two recurring prediction patterns plus one
configuration-specific parsing failure:

\textit{(a) Catch-all over-prediction.} In 7 of 8 LLM variants, the
destination in the most frequent wrong class pair is \emph{Miscellaneous};
in the remaining one (Haiku few-shot) it is \emph{Access}. Short, generic tickets also default to these
catch-alls (e.g.\ Llama few-shot routes 72 true-\emph{Access}
tickets to \emph{Miscellaneous}, such as ``password hello please
help passwords guest thank'').

\textit{(b) Surface-keyword bias.} LLMs latch onto single tokens
without context. Llama zero-shot routes 80 true-\emph{Access}
tickets to \emph{Hardware} (e.g.\ ``visual studio professional
license''), and Qwen zero-shot routes 58 true-\emph{Hardware}
tickets to \emph{Access} (e.g.\ ``weekend access and for the event'').

\textit{(c) Parse failures} (Llama few-shot only). At the 64-token
output budget the model emits a conversational preamble that
exhausts the budget before any JSON is produced
(15.1\,\% parse-error rate). \S\ref{sec:results-ablation}
analyzes this prompt/output-budget sensitivity. Because the LLM
scores rank errors weakly, the conservative LLM policy avoids most
of these errors mainly by deferring nearly all traffic; we do not
claim that modes (a)--(b) are reliably assigned low scores.

\section{Discussion}
\label{sec:discussion}

\subsection{Cost-aware abstention is a guardrail, not a fix}
Under the evaluated data and $(0,1,5)$ costs, thresholds selected
on calibration data reduce modeled cost for all four trained
classifiers. The fixed
break-even rule instead defers all traffic for seven prompted LLM
conditions. This does not improve those classifiers; it prevents
their weak or non-discriminative confidence signals from becoming
automatic routing errors.

ECE and discrimination answer different questions. Isotonic can
correct average probability levels without consistently improving
AUROC or AURC. Agreement can add ranking signal, but the stronger mixed
pool imports supervised information, requires multiple predictor
evaluations, and increases cost and latency.
The margin and entropy comparators show that the cost result is not
specific to one rejection score.
The post-hoc cost grid further shows that the qualitative benefit is
not confined to $(0,1,5)$, while its one slight held-out reversal
reinforces the need to freeze policy choices before test evaluation.

\subsection{Implications for deployment}
The framework could be integrated into an IT service management
pipeline as a deferral layer, subject to operational validation. Here,
``readiness'' remains conditional: an operator must estimate local costs,
reserve calibration data, freeze the policy, and monitor drift.
Defer-all is a warning under those inputs, not a permanent verdict
about a model family. The framework is agnostic to how the
underlying classifier was produced.

\subsection{Limitations and threats to validity}
\emph{Internal:} label noise, prompt/output-budget sensitivity, and
remote-endpoint change remain possible despite temperature 0. The
LLMs saw names and at most three examples per class, not label
definitions or fine-tuning; the observed accuracy gap therefore
mixes model behavior with unequal taxonomy access. A definition-rich
prompt is an important future control, but rerunning changed
proprietary endpoints would not be directly comparable to the frozen
May 2026 outputs.
\emph{External:} generalization to private enterprise corpora is
not established, and the cost model is illustrative: real cost
parameters need to be estimated per organization. The post-hoc grid
broadens the tested ratios but does not validate them as realistic
for any service desk. All results use one
public corpus, and remote endpoints are snapshots rather than a
longitudinal test; we do not test temporal drift, cross-organization transfer, or input
distribution shift, and multi-corpus validation is future work. A
single stratified random split can also place near-duplicate or
template-like tickets in both train and test, which could favor the
lexical classifiers. We normalized ticket text and represented it
with TF-IDF unigrams and bigrams capped at 30{,}000 features, defining
near-twins by cosine similarity $\geq 0.90$. No normalized texts are
exact duplicates; 3{,}986/47{,}837 tickets (8.3\,\%) have a near-twin,
and 674/9{,}567 random-split test tickets (7.1\,\%) have one in train.
As a robustness check, we connected each ticket to at most its five
nearest non-self neighbors above the same threshold, treated connected
components as groups, and formed a group-disjoint 33{,}379/14{,}458
train/test split. Its 69.8\,\% train share matches the primary split;
the remaining 30.2\,\% is a single test partition because this
classifier-accuracy check does not select an abstention threshold.
Because the finite-neighbor graph can omit qualifying edges, a direct
group-test-to-train scan still finds 26/14{,}458 near-twins (0.18\,\%).
Logistic-regression accuracy remains essentially unchanged
($0.847 \to 0.850$), suggesting that near-duplicate leakage does not
materially explain the classifier-versus-LLM gap.
Temporal splitting remains future work.
\emph{Construct:} accuracy on auto-routed tickets assumes a fixed
mapping from category to downstream queue, and we assume deferred
tickets are ultimately routed correctly, modeling human triage only
as a fixed cost $c_{\mathrm{triage}}$ (triager error and variable
handling time are out of scope). Misroute consequences can vary
across organizations and across ticket priorities.
We compare score-based rejection, including margin and entropy, but
not a jointly trained learning-to-defer system or a model of human
triager error. ECE uses ten fixed bins; AUROC and tie-group AURC are
included because low-resolution confidence can make ECE look good
without supporting selective routing.

\section{Conclusion}
\label{sec:conclusion}

We presented a reproducible empirical framework for asking whether
automatic ticket routing is cost-preferred under stated assumptions.
On one public corpus and primary illustrative $(0,1,5)$ costs, a
calibration-selected TF-IDF + LR policy reduces held-out modeled
cost by 38.1\,\%; in a post-hoc 16-setting cost analysis, frozen
calibration-selected policies beat both always-route and always-defer in 14
settings. The fixed break-even rule defers every ticket for
seven prompted LLM conditions and routes only 19/1{,}000 for Haiku
few-shot. Isotonic lowers ECE but generally does not improve
correctness ranking. The result is a scoped deployment warning, not
a universal ranking of model families. Separating prediction,
confidence, rejection, and cost makes each assumption auditable and
allows future datasets, prompts, learned deferral methods, and local
cost estimates to be evaluated under the same protocol.

\IEEEtriggeratref{12}
\bibliographystyle{IEEEtran}
\bibliography{references}

\end{document}